\documentclass[runningheads]{llncs}
\usepackage{graphicx,todonotes,amsmath}
\usepackage[caption=false,font=footnotesize]{subfig}
\usepackage{caption}
\usepackage{url}

\graphicspath{{figures/}}
\begin{document}
\title{Timage -- A Robust Time Series Classification Pipeline}
%
%
\author{Marc Wenninger*\inst{1} \and
Sebastian P. Bayerl*\inst{2} \and
Jochen Schmidt\inst{1} \and
Korbinian Riedhammer \inst{2} }
\authorrunning{M. Wenninger et al.}
%
\institute{
Rosenheim Technical University of Applied Sciences, Dep. of Computer Science, \and
Technische Hochschule Nürnberg Georg Simon Ohm, Dep. of Computer Science\\
\email{\{marc.wenninger,jochen.schmidt\}@th-rosenheim.de}\\
\email{\{sebastian.bayerl,korbinian.riedhammer\}@th-nuernberg.de}
}

\maketitle              
\begin{abstract}

Time series are series of values ordered by time.
This kind of data can be found in many real world settings.
Classifying time series is a difficult task and an active area of research.
This paper investigates the use of transfer learning in Deep Neural Networks and a 2D representation of time series known as Recurrence Plots.
In order to utilize the research done in the area of image classification, where Deep Neural Networks have achieved very good results, we use a Residual Neural Networks architecture known as ResNet.
As preprocessing of time series is a major part of every time series classification pipeline, the method proposed simplifies this step and requires only few parameters.
For the first time we propose a method for multi time series classification:
Training a single network to classify all datasets in the archive with one network.
We are among the first to evaluate the method on the latest 2018 release of the UCR archive, a well established time series classification benchmarking dataset.

\end{abstract}

\section{Introduction}\label{sc:intro}
Time series are N-dimensional signals ordered by time resulting in $N+1$ dimensions, hence all real world data recorded over time are a time series.
The sequence of data points allows one to identify higher features based on the combination of multiple data points in respect to the order.
All time series classification approaches try to identify these higher features to separate the classes based on them.
Time series classification is a hard problem that is not yet fully understood and numerous attempts have been made in the past to create generic and domain specific classification methods.
Because of the diverse domains where time series are present, the research and methods are diverse as well.
%
Many proposed methods for time series classification are actually classification pipelines: Multi-stage processes combining preprocessing steps and the actual classification method that separates the classes.
Preprocessing is used to provide more easily separable features, e.g., by reducing dimensionality, identifying regions of interest, or simply reducing length, always with the goal to support the classification algorithm.

During preprocessing, domain specific knowledge plays a vital role, as domain knowledge allows for identifying and removing noisy features and boost class-unique features by transferring knowledge from related research.
A generic classification pipeline needs to overcome the lack of domain knowledge. The authors of the UCR data set came to the conclusion that most probably there will never be one classification pipeline that rules them all, but every pipeline will have its advantages and disadvantages in the specific domain \cite{Bagnall2017}.
The reason for this lies within the differences of where higher features are located in the data. The importance of adaptable feature extraction is underlined by the success of classification pipelines that focus on feature extraction such as Weasel (Word ExtrAction for time SEriescLassification) \cite{weasel_2017} and BOSS (Bag-of-SFA-Symbols) \cite{Boss_2015}, which are able to adapt the feature extraction to different domains.
These methods preprocess the time series by encoding sliding windows into strings based on a similarity measure, followed by a classification using 1-nearest-neighbor. 
A nearly preprocess-less classification pipeline was proposed by \cite{UCR_LSTM_2018} using a Convolutional Neural Network (CNN) with a Long Short-Term Memory (LSTM) path.
\cite{UCR_CNN_2017} proposes a pipeline where time series are encoded as Reccurence Plot (RP) images, which are then classified using a CNN.
Encoding the time series of a 1D signal into a 2D image introduces new texture features, that allows one to utilize research efforts from image classification on time series classification.
RP in combination with a nearest neighbor classifier are used by \cite{ts_classification_recu_2015}.
The encoding of time series into images has also been used for sound classification, but instead of only using RP, \cite{music_instrument_park_2015} also evaluated the use of spectrogram images in combination with a CNN architecture that takes both image representations as the network input. \cite{enviromental_sounds_image_2017} evaluate the use of Spectrogram, Mel-Frequency Cepstral Coefficients (MFCC), and RP for the classification of environmental sound recordings using the AlexNet and GoogLeNet image recognition networks.
\cite{IsmailFawaz2018deep} presented results of an extensive experiment on feeding time series directly into 9 different well known neural network architectures such as MLP, FCN and ResNet. Their research shows that ResNet outperforms the others on the UCR archive.
The same author also showed that CNNs can benefit from transfer learning by reusing network weights \cite{IsmailFawaz2018transfer}.

Our contribution is a robust time series classification pipeline using transfer learning on ResNet with RP called \emph{timage}.
The ResNet network architecture is kept unchanged apart from adapting the input and output size of the network to fit the dataset.
As a starting point for transfer learning we reuse the network weights from the ImageNet challenge \cite{deep_residual_networks_for_image_resnet_152_2015} and further experimented with different strategies to obtain better weights.
We present a Single Classifier (SC) for each dataset in the UCR archive as done by all published models and are among the first to evaluate our method on the latest 2018 UCR release.
For the first time we present an All Classifier (AC), a single classification model trained on all datasets in the archive at once.
\section{Method}\label{sc:method}

\subsection{Approach}\label{ss:approach}

The main idea of this paper is to use existing knowledge in the form of pretrained image recognition networks to classify time series.
In order to take advantage of these neural networks that were trained on huge amounts of image data, time series must be converted to images.
This can be achieved by representations used for qualitative analysis like Recurrence Plots (RP), Spectrograms, Gramian Angular Summation/Difference Fields (GASF/GADF), and Markov Transition Fields (MTF) \cite{wang_image_ts_2015,recurrence_plots_marwan_2007,digital_signal_alessio2016}.
For our experiments we focus on RP for time series representation as proposed in \cite{ts_classification_recu_2015,music_instrument_park_2015}.
\subsubsection{Transfer Learning}

One of the main concepts exploited in this paper is called transfer learning.
The general assumption behind it is that something learned for one problem can be used to improve generalization for another problem.
It is assumed that many factors which explain variations in the first problem are also relevant for a similar problem.
Simply speaking, knowledge created by solving the first problem is used and applied to solve another problem  \cite{goodfellow-et-al-2016}.
This is especially useful when only a small amount of training data is available.
In the case of the UCR archive, this means that even the sets with only very little training samples can be properly learned by a neural network which usually needs huge amounts of training data to converge.

As we wanted to explore the concept of transfer learning even further, we tried different strategies to obtain new weights.
The first approach was ultimately the most successful one using the original weights obtained by training the networks on the ImageNet dataset and then use these weights for our networks.
A second method to obtain better weights for transfer learning was to train an AC system and then use these weights in training SC systems.
Another approach was training an AC not to separate all different classes but to separate datasets.
For this, the UCR archive data was relabeled to only have one label per dataset.
This means the network is trained to separate the datasets and not the classes within each dataset.
This approach in itself worked reasonably well but using these weighs for training AC or SC systems did not improve overall accuracy.
An interesting observation that could be made was that AC and SC systems trained on weights obtained by training with the RPs converged faster than systems using the original ResNet weights. 

\subsubsection{Reccurrence Plots}

The motivation for RP is that recurrence is a property of many natural and real world systems.
Situations or states that have been observed once are often followed by similar states.
RP can be used to visualize these recurrences \cite{recurrence_plots_marwan_2007}.
RP plot the recurrence matrix that is formally defined by equation \eqref{eq:recurrence}, where $N$ is the number of measured points \begin{math} \vec{x_i}, \ \epsilon \end{math} a threshold and \begin{math}\Theta\end{math} the Heaviside or step function where \begin{math} \Theta(x) = 0 \end{math} if \begin{math} x \le 0\end{math} and \begin{math} \Theta(x) = 1 \end{math} otherwise.
A proper distance measure has to be chosen for \begin{math}||.|| \end{math}, such as Euclidean distance or cosine distance.
\begin{equation}\label{eq:recurrence}
  R_{i,j} (\epsilon) = \Theta(\epsilon - ||\vec{x_i} - \vec{x_j}||), i, j = 1,...,N
\end{equation}
For states that are in an \begin{math} \epsilon \end{math}-neighborhood, the notion in \eqref{eq:notion} can be used:
\begin{equation}\label{eq:notion}
  \vec{x_i} \approx \vec{x_j} \iff R_{i,j}\equiv 1
\end{equation}
The RP is then generated by binarizing the recurrence matrix using a threshold \cite{recurrence_plots_marwan_2007}.
These plots are particularly useful to humans in qualitative assessment because patterns can be quickly identified visually.
For the use in image classification, more data in the form of more discriminable values in between might be desireable.
This is the main reason for using a modification of recurrence plots where no thresholding is performed to get more information encoded into the image.
Input data usually comes in the form of feature vectors, and the pairwise distances between those are computed, which creates the distance matrix $D_{i,j}$:

\begin{equation}\label{eq:distanceplot}
  D_{i,j} = ||\vec{x} - \vec{y}|| \quad.
\end{equation}

These pairwise distances are then plotted.
This modification of the RP is also known as \emph{unthresholded recurrence plot} or, maybe more appropriate, \emph{distance plots} as described in \eqref{eq:distanceplot} \cite{recurrence_plots_marwan_2007}.
We chose to use Euclidean distance for our plots.
After calculating the distances we apply some normalization by using a Min-Max scaler and a threshold cut-off at three times the standard deviation of all distances in the distance matrix.
\eqref{eq:actualplot} represents the plots used in this paper best:

\begin{equation}
  D_{i,j} = D_{i,j} (d \leq 3\sigma) =
  \begin{cases}
  3\sigma & d \ge 3 \sigma \\
  d                  & d < 3 \sigma
  \end{cases} \quad.
  \label{eq:actualplot}
\end{equation}

\begin{figure}[tb]
  \centering
\subfloat[][Class \emph{A} from the Adiac dataset]{
    \includegraphics[width=0.4\textwidth]{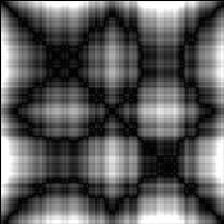}
    \label{fig:adiacA}
}
\subfloat[][Class \emph{B} from the Adiac dataset]{
    \includegraphics[width=0.4\textwidth]{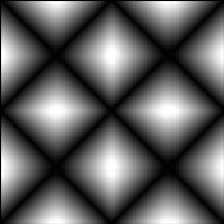}
    \label{fig:adiacbB}
    }
\caption{Distance Plots of two classes from the Adiac dataset, where classes represent the outlines of unicellular algae taken from images.}
\label{fig:adiacplot}
\end{figure}

The resulting plots show a gray-shaded pattern opposed to classical RP that have a simple black and white pattern.
Two samples taken from UCR archives Adiac dataset can be found in Fig.~\ref{fig:adiacplot}. 

Compared to methods like WEASEL or BOSS, our proposed method does not put much emphasis on extracting the best hyper-parameters for each dataset.
The only parameter required in its configuration is the image size.
By using the unthresholded RPs we got rid of the $\epsilon$ parameter used in traditional RPs which makes configuration easier and also leads to a plot that contains more information.
By using Euclidean distance by default and not trying to determine the best distance measure for each dataset we reduce configuration even more.
The individual datasets in the UCR archive are treated equally and no individual adaption is performed.
Within every experiment, the dataset is preprocessed in the same way, and the same resolution is used for all images in order to be as generic as possible.
It is reasonable to assume that this will lead to information loss, but through this inter-class normalization  we gain more easily comparable images of equal size.
To compensate for the in part heavily skewed datasets, we used class weights during training.

\subsection{Experimental Setup}\label{ss:setup}

Experiments were run on a HP Server with 1 TB of RAM, two Intel Xeon E5-2650(28 cores @ 2.60 GHz), and NVIDIA Tesla V100 graphics cards.
The server was running Ubuntu 16.04 LTS.
To improve reproducibility, Docker containers were used during training.
Docker encapsulates the runtime environment for the experiments. 
That way the environment is easy to recreate and version, as the runtime environment is simply described in structured text files.\footnote{\url{https://www.docker.com/}}
All neural networks used in this paper were implemented using the Keras API with a TensorFlow back-end \cite{keras_2015,tensorflow2015-whitepaper}. 
Keras also has a ready-to-go implementation of a Deep Residual Network (DSN) with 50 layers that will be referred to as ResNet-50 in this paper.
Also a DSN with 152 layer was used.\footnote{\url{https://github.com/flyyufelix/cnn_finetune/blob/master/resnet_152.py}}
Its implementation is based on code published on github.\footnote{\url{https://gist.github.com/previtus/c1a8604a4a07de680d5fb05cebfdf893}}
It will be referred to as ResNet-152.
The networks were trained with categorical cross entropy as the loss function, which is widely used for multi class problems, and used a Stochastic Gradient Descent (SGD) optimizer. 
The results were evaluated using the models' accuracy.
\section{Data}\label{sc:data}
The datasets used in this paper are all taken from the UCR Time Series Archive that was first introduced in 2002 and since then has been used in more than one thousand publications.
The archive got an update in 2018, expanding it from 85 to 128 datasets.
The new datasets on average have a higher number of training samples and also contain sets with variable length time series to represent many real-world problems \cite{ucr_2018}, ranging from the movement of insect wings to the energy consumption profile of electronic devices.
Most datasets in the archive have already been Z-score normalized.
For the sets that were not yet normalized, Z-score normalization was applied.
The datasets are split in fixed train and test sets.
The authors of the archive argue that the fixed train/test splits lead to more reproducible results and make publications and methods easier to compare.
The sizes of the individual datasets vary significantly and in some of the sets, the class distribution is strongly skewed and class distributions of test and train set also differ.
For example, the Chinatown train set consists of only 20 samples for 2 classes with 10 samples each, but the test set contains 95 samples of one class and 250 of the second class.

Another dataset that has at least indirectly been used in this paper is ImageNet, an image database containing millions of pictures\footnote{\url{http://image-net.org/}}.
The publishers of the dataset hold an annual competition, the Large Scale Visual Recognition Challenge (ILSVRC).
The challenge was won with a Deep Residual network as proposed in \cite{deep_residual_networks_for_image_resnet_152_2015}.
The network together with weights obtained by training on the ImageNet dataset was published and is widely used in transfer learning applications.\footnote{\url{https://keras.io/applications/#resnet50}}
\footnote{\url{https://www.mathworks.com/help/deeplearning/ref/resnet50.html}}
\footnote{\url{https://pytorch.org/docs/0.4.0/_modules/torchvision/models/resnet.html}}
\section{Experiments}\label{sc:experiments}
To the best of our knowledge, there have been no publications using the complete UCR 2018 archive at the time of writing of this paper.
Some results on the new dataset were only recently presented by \cite{insights_lstm_ucr_2019}.
Therefore, we present our results on the UCR 2018 archive separated into two tables.
The datasets already contained in the 2015 archive are compared with results from relevant publications and the new datasets are compared to the Dynamic Time Warping (DTW) published on the archives website\footnote{\url{https://www.cs.ucr.edu/~eamonn/time_series_data_2018}}$^{,}$ \footnote{Online results at \url{https://github.com/patientzero/timage-icann2019}}.


In total we ran 3942 experiments during our work on this paper.
Detailed results of all the experiments performed can not be properly presented nor discussed.
We want to primarily present two methods.
They will be referred to as All Classifier (AC) and Single Classifier (SC).
The AC was trained on all images from the UCR archive at once with its 1118 overall classes, i.e. the AC network has 1118 targets using one-hot encoding.
The SC is the classical approach to train one DNN per dataset.
The AC was evaluated in the same way as the SC, using only one dataset at a time, resulting in individual accuracy for each dataset. 
Final results are presented for gray-scale images with a resolution of 224x224, the resolution used in the original ResNet publication.
We experimented with different higher resolutions up to 512x512, but this did not improve classification accuracy significantly.
For some data in the UCR archive this means length reduction, performed using average pooling after calculating the RP matrix.
We also tried non gray-scale pictures with three colors channels by mapping the distance matrix to a color scheme, which  lead to overall worse results.
Detailed results on the 2015 datasets are shown in Table \ref{tbl:results-2015}, results on 2018 datasets in Table \ref{tbl:results-2018}.

The AC did exceed our expectations of it being nothing but an interesting idea by far.
Results for some datasets are very good and come close to state of the art systems and results from the SC, but for some datasets it completely fails, being worse than the default rate.
The default rate is defined by the accuracy a classifier would achieve by always predicting the most probable class.
We need to keep in mind that using a single classifier for all datasets within the UCR archive increases the complexity dramatically since it requires separating the datasets as well.
Maybe hyper-parameter tuning with special regard to the huge amount of classes might lead to better overall results.

\begin{figure}[tb]
    \centering
      \subfloat[][CricketX]{
          \includegraphics[width=0.33\textwidth]{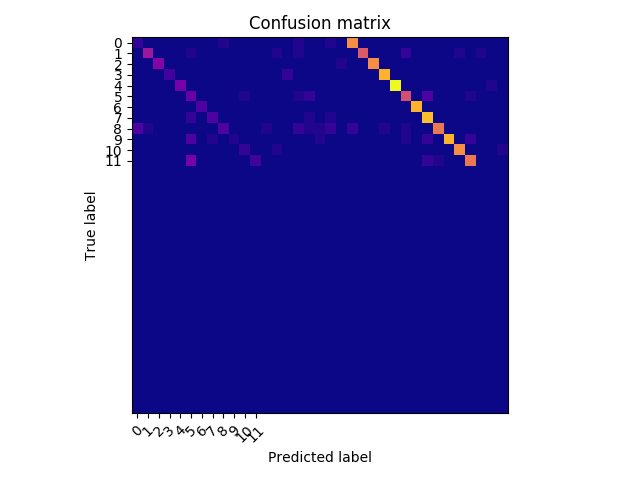}
          \label{fig:cricketX}
          }
      \subfloat[][CricketY]{
          \includegraphics[width=0.33\textwidth]{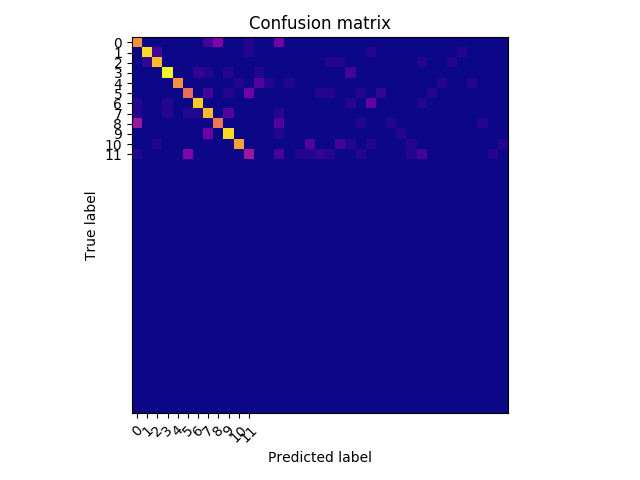}
          \label{fig:cricketY}
          }
      \subfloat[][CricketZ]{
          \includegraphics[width=0.33\textwidth]{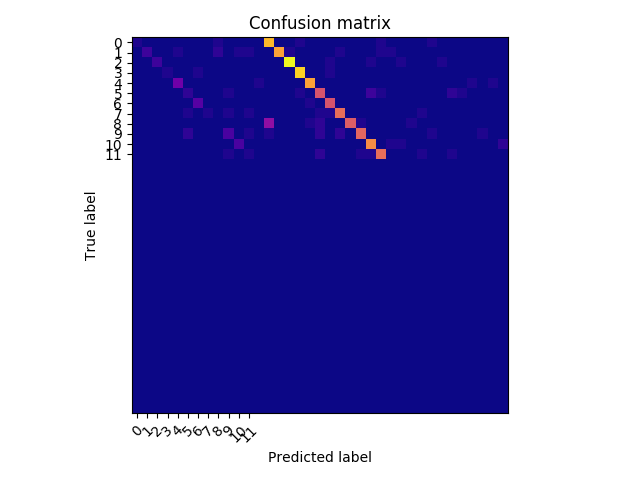}
          \label{fig:cricketZ}
          }
    \caption{Confusion Matrix for AC on the Cricket datasets, which are accelerometer data taken from actors performing Cricket gestures.}
    \label{fig:confusionmatrix}
\end{figure}

Detailed results confirm some of the things we know about recurrence plots and the conclusions that can be drawn about a time series by looking at the plot.
Patterns found in the images by the DNN will probably be too similar to be differentiated properly.
An in-depth look at misclassifications revealed that some of the very poor results were almost completely misclassified within a single other dataset.
An example of such a plot can be found in Fig. \ref{fig:confusionmatrix}.
The datasets are very similar in general.
An accuracy of 64\% could be reached with both AC classifiers, which is not particularly good but still better than the baseline (DTW) published by the authors.
The AC completely fails to classify anything correctly within CricketX and CricketZ.
Both datasets were almost exclusively classified as classes belonging to the CricketY dataset.
A probable explanation for this can be found in shuffling and batching of training data.
Shuffling the training data before feeding it in batches to the network may lead in case of very similar classes to over prioritizing earlier seen classes, thus failing to learn the classes later seen.
In other random states or in a different order the \mbox{(mis-)classification} results might be reversed.

Since there has been more research done on SC classifiers on the UCR archive we also present results in detail.
Table~\ref{tbl:accdiff} shows the relative accuracy differences of the SC timage compared to state of the art systems.
Table~\ref{tbl:accdiff}(a) shows the differences for the UCR 2015 archive compared to LSTM and WEASLE \cite{UCR_LSTM_2018,weasel_2017} and Table~\ref{tbl:accdiff}(b) for the new datasets in the UCR 2018 archive compared to the baseline.
For the UCR 2015 \emph{timage} outperforms WEASEL using ResNet-50 for 41 datasets, and using ResNet-152 for 37 datasets.
Same accuracy is achieved for 5 datasets for both ResNet-50 and ResNet-152.
Compared to the LSTM, \emph{timage} performs better for 7 datasets using ResNet-50 and for 8 datasets using ResNet-152.
Equal results were achieved in 7 cases using ResNet-50 and 4 using ResNet-152.
For 18 datasets relative classification accuracy was still comparable and only within a 5\% relative difference using ResNet-50 compared to WEASEL and 22 for ResNet-152.
Respectively for LSTM, 30 are within the 5\% range using ResNet-50 and 28 using ResNet-152.
For 8 datasets from the UCR 2015 archive, timage could achieve the best classification accuracy so far as shown in Table \ref{tbl:results-2015}.
Comparing the results for the UCR 2018 archive with the baseline, timage outperforms it in 26 out of 41 cases using ResNet-50 and 27 using ResNet-152.
For 5 datasets results were within the 5\% range using ResNet-50 and 3 using ResNet-152.

The results that can be achieved with the SC are comparable to state of the art systems as shown by Table \ref{tbl:results-2015}.
The method is able to classify all datasets well with the exception of ShapeletSim and TwoPatterns, where it is still better than the default rate.

\begin{table}[tb]
    \caption{Histograms of relative accuracy changes of SC models compared to Weasel and LSTM on (a) UCR 2015 and baseline (DTW) on (b) UCR 2018.\newline}
    \parbox{.5\linewidth}{
        \centering
        \begin{tabular}{ccc}
            \includegraphics{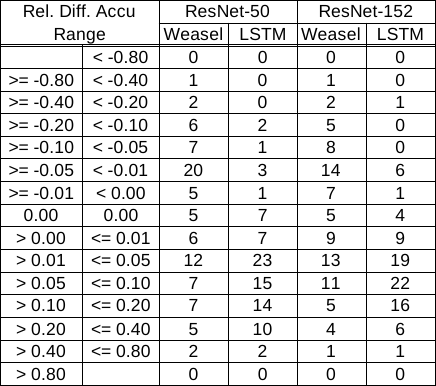}
        \end{tabular}
        \caption*{(a)}
        }
    \parbox{.5\linewidth}{
            \centering
            \begin{tabular}{ccc}
                \includegraphics{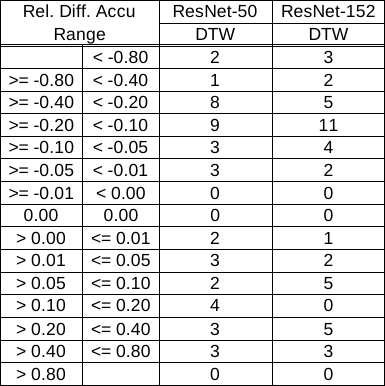}
            \end{tabular}
            \caption*{(b)}
    }
    \label{tbl:accdiff}
\end{table}

\begin{table}[tb]
    \centering
    \caption{Accuracy of experiments on UCR 2015 archive. Best results are highlighted.\newline}
    \includegraphics[scale=0.83]{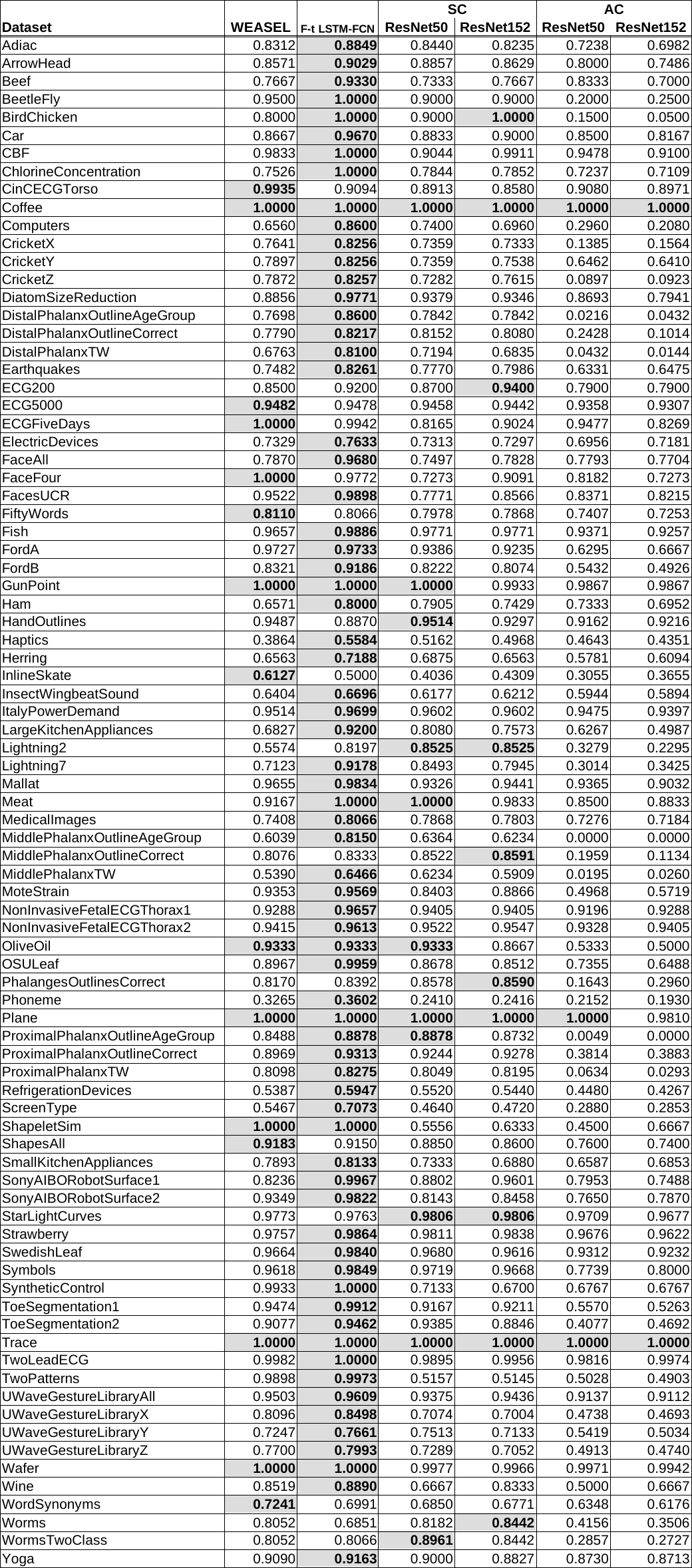}
    \label{tbl:results-2015}
\end{table}

\begin{table}[tb]
    \centering
    \caption{Accuracy of experiments on new entries of UCR 2018 archive. Best results are highlighted.\newline}
    \includegraphics[scale=0.83]{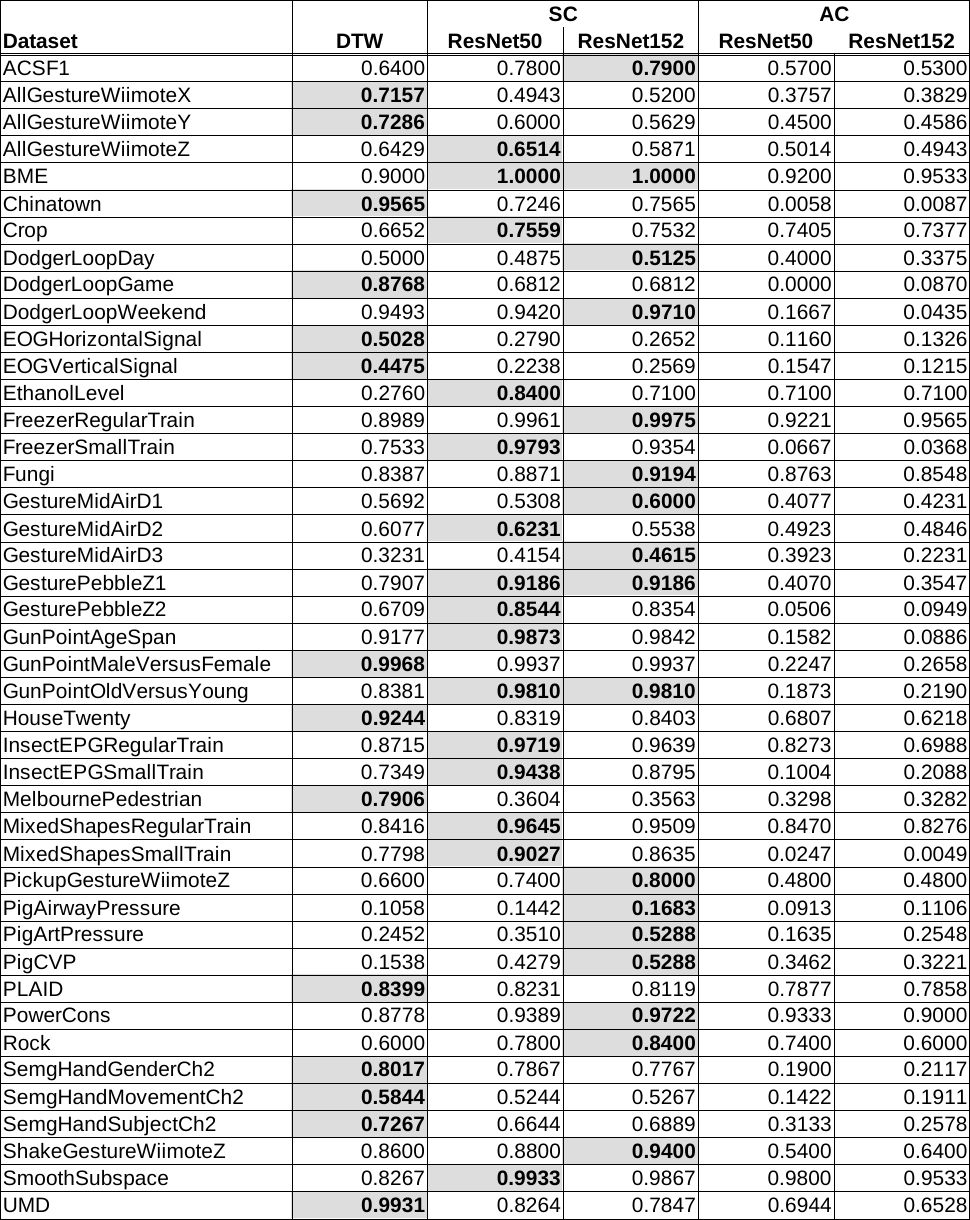}
    \label{tbl:results-2018}
\end{table}

\clearpage
\section{Conclusion}\label{sc:conclusion}

Especially in the light of the huge diversity of datasets in the UCR archive, the robustness of using transfer learning on ResNet to classify time series by using RP is surprising.
A number of changes in the preprocessing step like increasing the resolution or using a color scheme did not help the DNN to separate the classes more accurately compared to our initial experiments.
Different strategies to obtain weights for transfer learning did not lead to significant improvements but had impact on the time networks needed to converge.
The classification pipeline presented generally simplifies what needs to be done to detect patterns and correctly classify time series.

The analysis of confusion matrices for datasets that performed extremely bad with the AC show that some of them almost exclusively misclassified within one other dataset.
They do not fail the task of separating classes within one dataset, but fail on the harder problem of an AC, that is separating not only classes within a dataset, but to separate very similar datasets as well.
This supports some of the findings regarding RP for time series, and the conclusions regarding its properties, that can be drawn from looking at the respective RP: Similar time series produce very similar patterns \cite{recurrence_plots_marwan_2007}.\footnote{\url{http://www.recurrence-plot.tk/}}
Thus failing to always boost more easily separable features.

The nature of the UCR archive simplifies the classification task, because due to mostly fixed length and nearly perfect alignment of time series, some difficulties, like segmentation, faced in real world scenarios are removed.
In the case of the UCR archive all data is pre-segmented, which in itself is a very hard task to begin with.
Some of the harder tasks are included in the UCR 2018 archive where some datasets contain time series of different length and overall classification results are not as good as for the UCR 2015 archive. 
In the future it will be interesting to adapt and evaluate \emph{timage} to real world scenarios where, for example, a sliding window is used.

\section*{Acknowledgement}
This work is supported by a research grant of the Bayerisches Staatsministerium für Bildung und Kultus, Wissenschaft und Kunst and further by the Bayerische Wissenschaftsforum (BayWISS).
%
%
%
\bibliographystyle{splncs04}
\bibliography{literatur}

\end{document}